\RequirePackage{fix-cm}
\documentclass[twocolumn]{svjour3}          
\smartqed  
\usepackage{times}
\usepackage{epsfig}
\usepackage{amssymb}
\usepackage{siunitx}
\usepackage{diagbox}
\usepackage{array}
\usepackage[colorlinks,citecolor=blue]{hyperref}

\usepackage{graphicx}

\usepackage{natbib}
\usepackage{amsmath}
\usepackage{multirow}
\usepackage{colortbl}
\usepackage[T1]{fontenc}
\usepackage{graphicx}

\usepackage[colorlinks,citecolor=blue]{hyperref}
\usepackage{xcolor}
\usepackage{subcaption}
\usepackage{booktabs}
\usepackage{float}
\usepackage{tabularray}
\usepackage[ruled,vlined,linesnumbered]{algorithm2e}
\usepackage{amsmath} 
\usepackage{multirow}

\newcommand{\bst}{\color[HTML]{C00000}} 
\newcommand{\scd}{\color[HTML]{00B500}}
\newcommand{\ie}{ \textit{i.e.}}
\newcommand{\eg}{\textit{e.g.}}

\begin{document}
\begin{sloppypar}

\title{Harmonizing and Merging Source Models for CLIP-based Domain Generalization}


\author{Yuhe Ding\and Jian Liang \and Bo Jiang*\and Zi Wang\and Aihua Zheng\and Bin Luo*}
\institute{Yuhe Ding, Bo Jiang, and Bin Luo are with the School of Computer Science and Technology, Anhui University \email{madao3c@foxmail.com, jiangbo@ahu.edu.cn, luobin@ahu.edu.cn};   \\
 Jian Liang is with the Institute of Automation, Chinese Academy of Sciences. \email{liangjian92@gmail.com}; \\
 Zi Wang is with the School of Biomedical Engineering, Anhui Medical University. \email{ziwang1121@foxmail.com}; \\
 Aihua Zheng is with the School of Artificial Intelligence, Anhui University. \email{ahzheng214@foxmail.com} \\
 *Corresponding authors.
}

\date{Received: date / Accepted: date}
\maketitle

\begin{abstract}
CLIP-based domain generalization aims to improve model generalization to unseen domains by leveraging the powerful zero-shot classification capabilities of CLIP and multiple source datasets.
Existing methods typically train a single model across multiple source domains to capture domain-shared information. 
However, this paradigm inherently suffers from two types of conflicts:
1) sample conflicts, arising from noisy samples and extreme domain shifts among sources;
and 2) optimization conflicts, stemming from competition and trade-offs during multi-source training.
Both hinder the generalization, lead to suboptimal solutions.
Recent studies have shown that model merging can effectively mitigate the competition of multi-objective optimization and improve generalization performance.
Inspired by these findings, we propose Harmonizing and Merging (HAM), a novel source model merging framework for CLIP-based domain generalization.
During the training process of the source models, HAM enriches the source samples without conflicting samples, and harmonizes the update directions of all models.
Then a redundancy-aware historical model merging method is introduced to effectively integrate knowledge across all source models.
HAM comprehensively consolidates source domain information while enabling mutual enhancement among source models, ultimately yielding a final model with optimal generalization capabilities.
Extensive experiments on five widely-used benchmark datasets demonstrate the effectiveness of our approach, achieving state-of-the-art performance.

\keywords{Domain Generalization  \and Multi-modal Domain Generalization \and Model Merging}

\end{abstract}

\section{Introduction}\label{sec:intro}
Contrastive Language-Image Pretraining (CLIP) \citep{radford2021learning} is a powerful vision-language model trained on a vast dataset of image-text pairs. 
By aligning images and textual descriptions in a shared embedding space, CLIP enables strong zero-shot classification capabilities, making it well-suited for tasks that require adaptability to new domains.
However, the effectiveness of this capacity often degrades when there are unknown distribution shifts during testing.
Leveraging CLIP for domain generalization (CLIP-DG) \citep{yu2024clipceil,han2025alignclip}, which seeks to utilize labeled source domain data and pre-trained CLIP models to improve model performance under unknown domain shifts, has emerged as a promising research direction.

Existing approaches to CLIP-DG generally fall into two categories.
Some methods follow the classic domain generalization (DG) paradigm \citep{muandet2013domain}, aiming to extract representations that are highly correlated with classes while remaining invariant to domains \citep{yu2024clipceil, hu2024learn}.
Other methods \citep{li2024seeking, addepalli2024leveraging, bai2024soft, han2025alignclip} focus on leveraging the rich pre-trained knowledge within CLIP itself to enhance model robustness. 
Both approaches typically train a single model across multiple source domains.
However, this widely adopted one-size-fits-all paradigm presents inherent limitations that hinder generalization performance.
We identify two fundamental conflicts in this paradigm. 
The first is sample conflict, which arises from noisy samples and extreme domain shifts among different sources.
Noisy samples lead to instability in representation learning, while extreme domain shifts result in minimal or even non-existent shared information across domains, rendering effective learning difficult.
The second is optimization conflict, stemming from the competition and trade-offs inherent in multi-source training.
Conflicting gradients from diverse domains often lead to suboptimal convergence, disproportionately sacrificing the performance of domains with fewer samples.
Additionally, noise or outliers in dominant domains can cause overfitting, further degrading generalization performance.

Model merging \citep{yang2024model} has emerged as a promising solution for mitigating these conflicts.
By consolidating multiple trained models into a single model, model merging has demonstrated strong potential in multi-objective and multi-task learning.
This paradigm effectively balances trade-offs between multiple objectives \citep{tang2024towards} while producing smoother parameter landscapes that enhance generalization \citep{izmailov2018averaging, cha2021swad, wortsman2022model}.
It is worth noting that, unlike ensemble learning \citep{dietterich2002ensemble}, model merging performs parameter-level integration rather than output-level combination. 
This yields a single unified model that requires only one inference pass, eliminating the need to store multiple source models during deployment.
Inspired by these insights, we introduce Harmonizing and Merging (HAM), a novel "learning-then-merging" framework designed to address the inherent conflicts in CLIP-DG.

HAM consists of two key phases: learning and merging, \ie, learning a vision encoder for each source domain and merging them for a final model.
The text encoder is frozen and shared by all the source vision encoders.
In the learning phase, a vision encoder with strong generalization capabilities and rich domain-specific information is fine-tuned for each source domain.
Additionally, potential conflicts that may arise during the merging phase are preemptively alleviated by aligning parameter update directions to prevent cancellation due to opposite signs of model update vectors \citep{ilharco2022editing, yadav2023ties}. 
The merging phase then focuses on efficiently integrating these models while mitigating interference from redundant parameters.
Specifically, HAM consists of three modules, \ie, sample conflict-aware adaptive source enrichment, optimization conflict-aware parameter alignment, and redundancy-aware historical model merging.
The source enrichment module dynamically determines an adaptive confidence threshold for each source domain, selectively extracting non-conflicting samples from other domains to enhance domain robustness while preventing failures due to insufficient or poor-quality samples in weaker domains.
Our parameter alignment module incorporates a directional regularization term during source model training, which penalizes deviations from the average update direction, to proactively minimize merging conflicts. 
Finally, we aggregate historical models within each source domain and trim redundant parameters to derive the final merged model with smoothed updates and reduced redundancy.

By comprehensively consolidating domain-specific and domain-invariant information, HAM significantly enhances generalization to unseen domains.
Extensive experiments on five widely used DG benchmarks demonstrate that HAM achieves state-of-the-art performance on these benchmarks. 
These results underscore the effectiveness of model merging as a powerful paradigm for DG, providing a scalable and robust solution for this task.

Our contributions can be summarized as follows:
\begin{itemize}
    \item We analyze the two inherent conflicts in domain generalization, and propose HAM, a novel "learning-then-merging" framework for the domain generalization problem with a strong potential.
    \item We propose a source enrichment module and a parameter alignment regularization to enhance the source model while harmonizing all the source models.
    \item We propose a redundancy-aware historical model merging module to integrate the domain information while filtering out redundant parameter updates.
    \item HAM outperforms existing methods on the five benchmark datasets, demonstrating its effectiveness in the CLIP-based domain generalization field. The competitive results on the non-CLIP backbone demonstrate the generality of HAM.
\end{itemize}

\section{Related Works} \label{sec:rela}

\subsection{Domain Generalization}
Domain generalization (DG) \citep{zhou2022domain} aims to learn a model from multiple source domains that generalizes well to unseen target domains. 
DG has been explored in various settings and downstream tasks \cite{chen2024domain, huo2024domain, wang2024csdg}.
We introduce related works from two perspectives: traditional deep learning and vision-language foundation models.

\noindent
\textbf{Traditional domain generalization.} 
Traditional DG approaches work for general deep learning models, we introduce several classical categories of DG:
1) Domain alignment or invariant representation learning \citep{muandet2013domain, li2018domain, li2018deep, shao2019multi, motiian2017unified, wang2021respecting}, which aims to minimize the difference among source domains for learning domain-invariant representations.
1) Data augmentation, which improves model robustness by introducing domain variations through techniques such as image transformation \citep{volpi2019addressing, otalora2019staining} and style transfer \citep{yue2019domain, zhou2023semi, kang2022style};
3) Meta-learning, where the model is trained in a way that mimics domain shifts to enhance generalization \citep{li2018learning, liu2020shape, li2019episodic, zhao2021learning}; 
4) Ensemble learning, which aggregates multiple models trained on different source domains to enhance robustness \citep{zhou2021domain, ding2017deep, wang2020dofe}. 
This type is similar to us, where there is a domain-specific model for each source.
Such methods require integrating the output or prediction of the model, and need to save multiple source models simultaneously during inference, while we merge the source models into one model at the parameter level.
In addition, there are many other ways to solve the DG problem from other perspectives, such as self-supervised learning \citep{bucci2021self}, disentanglement representations \citep{piratla2020efficient}, robust representation learning \citep{cha2021swad, li2024seeking}, and reinforcement learning \citep{yarats2021improving}, etc.

\noindent
\textbf{CLIP-based domain generalization.}
In the field of DG, a significant body of recent work has emerged, focusing on developing DG methods specifically tailored for vision-language models (VLMs), particularly those built on the foundation of CLIP \citep{radford2021learning}.
Next, we introduce several typical works.
VLV2 \citep{addepalli2024leveraging} aligns and distills black-box VLM representations to student models while retaining their OOD generalization capabilities.
SPG \citep{bai2024soft} proposes a generative approach to create diverse soft prompts for VLMs using domain-specific knowledge, achieving SOTA on five DG benchmarks without manual prompt engineering.
PEGO \citep{hu2024learn} injects multiple LoRA \citep{hu2022lora} modules with orthogonal regularization into vision transformers to preserve pre-trained features while learning diverse domain-invariant knowledge.
CLIPCEIL \citep{yu2024clipceil} refines the feature channels in the visual domain to ensure they contain domain-invariant and class-relevant features by minimizing the inter-domain variance while maximizing the inter-class variance.
AlignCLIP \citep{han2025alignclip} addresses attention and predictive category misalignments in CLIP via attention alignment loss and semantic label smoothing, enhancing stability for domain shifts and unseen classes.
SFT \citep{li2024seeking} proposes a self-feedback training framework that iteratively refines loss landscapes to find consistent flat minima shared across domains.

\subsection{Model Merging}
Model merging \citep{yang2024model} aims to integrate the parameters of multiple individual models with distinct capabilities, creating a powerful model without requiring significant computational resources or access to the original training data of each model.
Unlike ensemble learning \citep{dietterich2002ensemble}, which retains all individual models and fuses their outputs during inference, model merging integrates models at the parameter level, resulting in a single final model for inference.
This property facilitates practical deployment.
Model merging can be applied to a variety of foundation models, including large language models \citep{akiba2025evolutionary}, multimodal large language models \citep{aiello2023jointly}, and visual generative models \citep{biggs2024diffusion}.
Moreover, model merging also arises in different machine learning subfields, especially in multi-task and multi-objective optimization.
In multi-task learning, directly merging models trained on diverse tasks can result in a single model capable of handling multiple tasks.
Task Arithmetic \citep{ilharco2022editing} proposes manipulating task vectors via arithmetic operations like negation, addition, and analogies to edit model behavior efficiently, enabling multi-task enhancement or bias mitigation without retraining.
Ties-Merging \citep{yadav2023ties} introduces a three-step method (Trim, Elect Sign, Disjoint Merge) to resolve parameter conflicts during model merging, significantly improving multi-task performance by eliminating interference from opposing weight updates.
AdaMerging \citep{yang2023adamerging} learns layer-wise or task-wise merging coefficients autonomously using entropy minimization on unlabeled data.
Multi-objective optimization involves finding a trade-off among the multiple objectives, which corresponds to identifying a set of Pareto optimal solutions.
Tang et al. \citep{tang2024towards} propose approximating the entire Pareto set by model merging.
They train an independent model for each objective and learn a routing network to balance the trade-offs between the multiple objectives. 
The input of the routing network is the task preference vector, and its output consists of the merging coefficients for the independent models. 

Model merging also improves the out-of-distribution generalization.
Classical methods merge weights obtained along one training trajectory.
Stochastic weight averaging (SWA) \citep{izmailov2018averaging} proposes averaging multiple points along the trajectory of stochastic gradient descend iterates to find a flatter minimum in the loss landscape, improving generalization without extra computational cost compared to traditional training.
WiSE fine-tuning \citep{wortsman2022robust} combines zero-shot and fine-tuned model weights via linear interpolation to enhance robustness against distribution shifts while preserving original model capabilities.
Furthermore, several studies have demonstrated that integrating multiple independently fine-tuned or trained models can significantly enhance generalization performance.
Model Soup \citep{wortsman2022model} proposes averaging weights of multiple models fine-tuned with different hyperparameters into a soup, where greedy soup—sequentially adding models that improve validation performance.
Model-Ratatouille \citep{rame2023model} recycles diverse auxiliary task fine-tunings of a foundation model by repurposing their weights as parallel initializations for target task training, then averaging them to enhance OOD generalization via feature diversity.
Existing model merging-based domain generalization methods typically merge weights along the training steps.
SWAD \citep{cha2021swad} finds flatter minima and suffers less from overfitting than does the vanilla SWA \citep{izmailov2018averaging} by a dense and overfit-aware stochastic weight sampling strategy.
While existing methods primarily focus on parameter smoothing to enhance model robustness, they largely overlook the inherent inter-domain conflicts. 
We address this critical limitation through merging models across training steps and domains simultaneously. 
This dual-aspect merging not only improves generalization performance but also mitigates domain conflicts.
\begin{figure*}
    \centering
    \includegraphics[width=.8\linewidth]{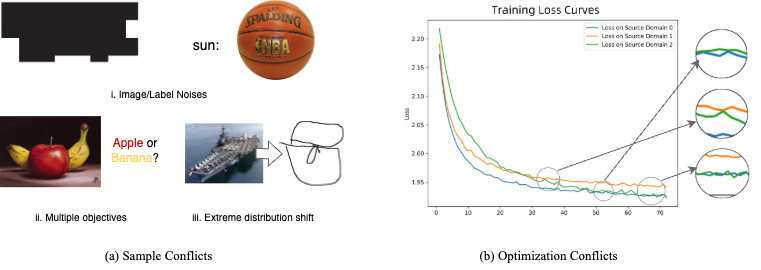}
    \caption{
    Illustration of dual conflicts in domain generalization:
    (a) Sample conflicts arise from four distinct types of data corruptions that can significantly degrade model performance.
    (b) Optimization conflicts emerge as an inherent challenge in multi-objective optimization, manifesting as competing gradient directions during training.
    }
    \label{fig:conflict}
\end{figure*}

\section{Method} \label{sec:mtd}
\subsection{Problem Setup} \label{sec:mth:setup}
\noindent
\textbf{Classification paradigm for CLIP.}
We denote the CLIP as $f = \{\phi,\xi\}$, where $\phi$ and $\xi$ notate the visual encoder and textual encoder, respectively;
$\mathcal{X}=\{x_i\}_{i=1}^{N}$ denotes the test dataset, where $N$ is the number of images.
$C=\{c_k\}_{k=1}^K$ represents the class names, \ie, label space, where $K$ is the number of classes. 
Classification of CLIP involves encoding both image and text prompts (\textit{e.g.}, "a photo of a \{class name\}") into embeddings. 
An image is classified by selecting the class whose textual embedding has the highest cosine similarity to the image's visual embedding,
\begin{equation}\label{eq:clip}
     \hat{y_i} = \operatorname{argmax}\limits_k(\operatorname{cos}(\mathbf{I}_k, \mathbf{V}_i)),
\end{equation}
where $\operatorname{cos}(\cdot)$ is the cosine similarity, $\mathbf{I}_k\in \mathbb{R}^{D\times K}$ is the text embedding of the prompt of the class name $c_k$; $\mathbf{V}_i \in \mathbb{R}^{D}$ denotes the visual embedding, $D$ is the dimension of embeddings.
%

\noindent
\textbf{Domain generalization based on CLIP.}
In this setting, a pre-trained CLIP model $f=\{\phi,\xi\}$ and $N$ labeled source datasets $\{\mathcal{X}_i\}_{i=1}^{N_S}$ are provided, and target dataset is unknown.
These source datasets, \ie, source domains, are distinct but related, known as domain shift.
Both these source datasets and the unknown target dataset share a common label space, encompassing $K$ classes.
Our goal is to learn generalizable information from the provided source datasets, thereby creating a robust CLIP model capable of accurately inferring the labels of an unknown target dataset.

\subsection{Motivation}

Existing methods in the community primarily follow two paradigms.
1) Domain-invariant representation learning \citep{yu2024clipceil, hu2024learn}, which seeks to identify and eliminate domain-specific information while preserving domain-invariant representations, thereby extracting transferable knowledge that remains consistent across diverse domains.
2) Robust representation learning \citep{li2024seeking, addepalli2024leveraging, bai2024soft, han2025alignclip}, which consolidates multiple source domains into a single domain, employing various regularization techniques to enhance model generalization and develop robust out-of-distribution performance.
While both approaches seek to learn a highly generalizable model directly, they inherently induce unavoidable inter-domain conflicts.
These two approaches share the fundamental objective of learning a maximally generalizable model, which can be formally cast as a multi-objective optimization framework with each source domain's risk constituting a separate objective. 
Such formulation intrinsically induces competition and conflicts among the competing objectives.

Recently, Tang et al. \citep{tang2024towards} provide evidence that parameter merging of multiple expert models can effectively capture inter-objective trade-offs, with each expert model specializing in distinct optimization objectives. 
Besides, a growing body of research \citep{izmailov2018averaging, cha2021swad, wortsman2022model} further demonstrates that model merging successfully integrates complementary knowledge across models while simultaneously enhancing generalization performance. 
Both aspects exhibit strong alignment with the key challenges of the domain generalization problem.
A basic framework is naturally induced, which trains a separate model for each source domain and merges them into a final DG model.

\subsection{Dual Conflicts in Domain Generalization.}
We conduct a deeper analysis of two types of inter-domain conflicts in DG problems, with an intuitive illustration provided in Fig. \ref{fig:conflict}.

The first type is \textbf{sample conflict}, shown in Fig. \ref{fig:conflict} (a).
In multi-source domain collaborative training, while domains may benefit from shared information through cross-domain knowledge transfer, this potential advantage is not universally guaranteed. 
We list three predominant forms of sample-level conflicts that can arise in this setting:
i) Image/Label Noise: External domains often comprise wild datasets where neither the presence of task-relevant visual information nor annotation accuracy can be guaranteed.
ii) Multi-object interference: A single image may contain multiple targets belonging to different categories, while single-label classification typically only assigns one dominant label. This frequently leads to model bias toward certain classes while neglecting others.
iii) Extreme domain shift: When the distributional discrepancy between source domains becomes excessively large, their shared information diminishes significantly, rendering samples from other domains practically uninformative for the learning process.

Another type is \textbf{optimization conflict}.
We show the training loss curves on each source domain of empirical risk minimization (ERM) on the TerraIncognita \citep{beery2018recognition} dataset shown in Fig. \ref{fig:conflict} (b).
During the mid-to-late stages of training, we observe competitive dynamics among losses across different source domains. 
Specifically, the optimization process exhibits an antagonistic pattern where a reduction in one domain's loss correlates with an increase in one or more other domains' losses. 
This phenomenon emerges as a fundamental characteristic of the multi-source optimization landscape.

\subsection{HAM: Harmonizing and Merging Source Models for CLIP-based Domain Generalization}
\begin{figure*}
    \centering
    \includegraphics[width=1\linewidth]{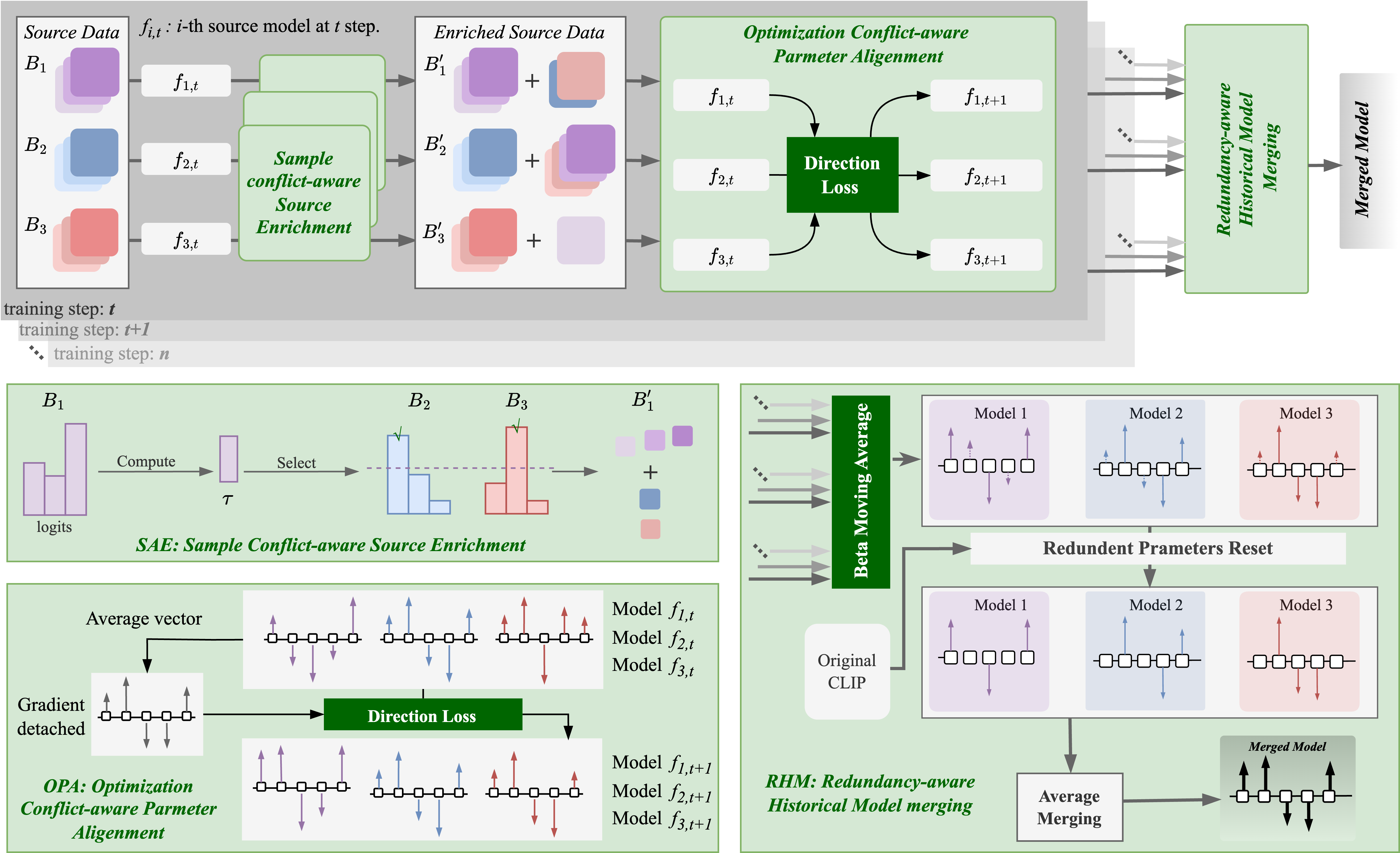}
    \caption{ The proposed framework for CLIP-based domain generalization. 
    It consists of three key modules: 1) Sample Conflict-aware Source Enrichment, which selects non-conflict out-of-distribution samples to enrich source domains; 
    2) Optimization-aware Parameter Alignment, which employs a direction loss to ensure consistent parameter updates across models; 
    and 3) Redundancy-aware Historical Model Merging, which trims and resets redundant parameters and merges models into a unified DG model.}
    \label{fig:pipeline}
\end{figure*}

In this section, we introduce our conflict-aware model merging method.
The proposed pipeline is shown in Fig. \ref{fig:pipeline}.
We first enhance each source $\mathcal{X}_i$ domain by leveraging data from other sources to train effective individual models. 
To address the inevitable sample conflicts where cross-domain samples may not always be beneficial, we introduce a sample conflict-aware adaptive source enrichment module. 
Subsequently, we fine-tune CLIP's visual encoder $f_i$ separately for each enriched source domain $\mathcal{X}'_i$ while keeping the text encoder fixed, then merge them into a single visual encoder. 
This isolated training approach temporarily avoids conflicts during the optimization process. 
However, this proves insufficient as parameter merging after training still introduces conflicts due to divergent optimization directions \citep{yadav2023ties}. 
To mitigate this, we propose an optimization conflict-aware parameter alignment module that adjusts each source model's parameter update directions to mitigate merging conflicts. 
Furthermore, recognizing that some conflicts remain unavoidable due to domain shifts, we develop a Redundancy-aware Historical Model Merging module that first integrates historical models from each source to obtain smoother parameters, then performs parameter trimming to eliminate redundancies, thereby minimizing merging conflicts to the greatest extent possible.
Next, we elaborate on these modules in detail.

\noindent
\textbf{Sample Conflict-aware Adaptive Source Enrichment (SAE).} \label{sec:mtd:sample}
While training domain-specific models using single-domain data is feasible for well-resourced domains, it becomes challenging for domains with limited samples, poor image quality, or noisy labels. 
Inspired by classical methods in semi-supervised learning \citep{berthelot2021adamatch}, we introduce an adaptive threshold $\tau_i$ that dynamically filters out conflicted samples from other domains, thereby enabling each domain to enrich its training data while mitigating sample conflict effects selectively.
This module is embedded before each training step, and the enrichment is conducted on the mini-batch level.
To be specific, at each training step, we denote the mini-batch data from the $i$-th source domain as $B_i = \{x_k^i, y_k^i\}_{i=1}^{n_i}$, where $n$ is the batch size. 
The adaptive threshold $\tau_i$ is the average confidence on $B_i$,
\begin{equation} \label{eq:tau}
    \tau_i = \frac{1}{n}\sum_{k=1}^n \operatorname{max}(f_i(x^i_k)),
\end{equation}
where $f_i(x)=\operatorname{softmax}(\operatorname{cos}(\phi(x),\mathbf{I}))$ is the probability distribution of sample $x$ on $i$-th model $f_i$, $\mathbf{I}$ is the textual embedding of class names, $max(f(x))\in(0,1]$ is the confidence---the maximum class probability---of $f(x)$.
Then the source enrichment can be formulated as,
\begin{equation} \label{eq:enrich}
\begin{split}
    & B'_i = B_i\cup (\bigcup_{j\neq i}\bar{B}_{j;i}), \\
    & \bar{B}_{j;i} =\bigcup_{k=1,j\neq i}^{n}\{(x^j_k,y^j_k)\in B_j\}\cdot\mathbb{I}(\operatorname{max}(f_i(x^j_k))>\tau_i),
\end{split}
\end{equation}
where $\mathbb{I}(\cdot)$ is the indicator function, to filter out those samples with confidence scores on $f_i$ below the threshold $\tau_i$.
$\tau_i$ quantifies the domain shift for samples originating from other domains.
The standard cross-entropy loss is employed on the enriched source data $B'_i$ to train the separate model,
\begin{equation}
\mathcal{L}_{CE} = -\frac{1}{{N_B}'} \sum_{k=1}^{{N_B}'} \sum_{c=1}^C \mathbb{I}(c = y_i) \log(f_i(x_k)[c]),\  x_k\in B'_i
\end{equation}
where $C$ is the total number of categories, ${N_B}'$ is the total number of samples in $B'_i$.
When samples exhibit inferior quality, contain multiple conflicting targets, or share minimal common information with the $i$-th domain, they are automatically assigned low confidence scores and filtered out, thereby effectively mitigating sample-level conflicts.
An adaptive threshold dynamically allocates more informative samples to resource-constrained domains, strengthening their basic performance.

\noindent
\textbf{Optimization Conflict-aware Parameter Alignment (OPA)} \label{sec:mtd:optim}
While direct optimization conflicts are avoided during training, these conflicts are not eliminated but rather shifted to the model merging stage.
We first give the definition of model update vector $\mathbf{v}=\theta-\theta_0$, which is computed by subtracting the pre-trained model’s parameter values $\theta_0$ from those of the fine-tuned model $\theta$.
In related works \citep{ilharco2022editing}, this concept is alternatively referred to as a "task vector". 
To avoid terminological ambiguity, we adopt the more proper designation of "model update vector" for our task.
When model update vectors contain parameters with opposing signs (\ie, conflicting directional updates), direct arithmetic averaging during merging may induce destructive interference, thereby compromising model performance \citep{yadav2023ties}.
As a result, we propose a novel regularization term across all layers of each model,
\begin{equation}\label{eq:sign}
\begin{split}
    \mathcal{L}_{sign} = \frac{1}{N_L}\sum_{l=1}^{N_L} \operatorname{max}(0, -\mathbf{v}^l_i\cdot \bar{\mathbf{v}}^l), 
\end{split}
\end{equation}
where $\mathbf{v}^l_i$ is the model update vector of $l$-th layer in $i$-th model, $\bar{\mathbf{v}}^l = \frac{1}{N_S}\sum_{i=1}^{N_S}\mathbf{v}^l_i$ is the average model update vector of $l$-th layer across all the source models, $N_S$ and $N_L$ denotes the the total number of source domains and model layers, respectively.
The gradient of $\bar{\mathbf{v}}^l$ is detached during training.
$\mathcal{L}_{sign}$ activates when $\mathbf{v}_i$ and $\bar{\mathbf{v}}$ exhibit opposing signs, enforcing consistent directional alignment between parameters during optimization.

\noindent
\textbf{Redundancy-aware Historical Model Merging (RHM).} \label{sec:mth:merge}
While parameter averaging across all source models is a fundamental model merging method, this naive aggregation strategy typically yields suboptimal performance.  
We introduce a dual-level merging approach that operates across both training steps and domains.
1) Historical model merging: we first perform historical model merging to smooth model update vectors and prevent potential overfitting in later training stages.
2) Domain-level merging: next, we apply domain-level merging to these intermediate source models.
This process is formally expressed as:
\begin{equation}\label{eq:merge1}
\begin{split}
    &\theta = \frac{1}{N_S}\sum_{i=1}^{N_S}\theta_i,\ \theta_{i} = \sum_{t=0}^{N_T}\frac{\gamma_t}{\sum_{k=0}^{N_T}\gamma_k}\cdot\theta_{i,t}, 
\end{split}
\end{equation}
where $N_T$ is the total number of training steps, $\gamma_t$ is the weight assigned to the model at step $t$.
Storing all intermediate models is computationally expensive. 
To improve efficiency, we employ a Moving Average strategy \citep{shu2023clipood, izmailov2018averaging},
\begin{equation}\label{eq:bma}
    \theta_{i,t} = \frac{\sum_{k=0}^{t-1}\gamma_k}{\sum_{k=0}^{t}\gamma_k}\cdot \theta_{i,t-1}+\frac{\gamma_k}{\sum_{k=0}^{t}\gamma_k}\cdot \theta_{i,t}.
\end{equation}
Note that we assign different weights to models at different training steps to balance retaining knowledge from the original CLIP while adapting to source domain information. 
Specifically, model weights across training steps are sampled from a Beta distribution $\gamma_t = \operatorname{Beta}(\beta,\beta)(\frac{t+0.5}{T+1})$, where $\beta$ is the hyperparameter.
The Beta distribution exhibits a U-shaped pattern, emphasizing models from both the early and late stages of training \citep{shu2023clipood}.
In contrast, models from different domains are given equal weights, as the target domain remains unknown, making it impossible to determine which source is more relevant.

Although historical model merging and the proposed sign loss partially mitigate parameter conflicts, they cannot fully resolve them due to inherent optimization constraints. 
Increasing the number of participating parameters exacerbates conflicts, so we further address this by reducing the number of parameters involved.
Empirically, model parameters exhibit significant redundancy \citep{hoefler2021sparsity, thimm1995evaluating}. 
Only the parameters with the largest magnitudes—representing the most substantial updates—are determinative for model performance. 
Leveraging this, we introduce a redundancy-aware model merging strategy, which takes the historically merged source models as inputs and outputs a model with reduced redundant updates.
First, we trim the parameters of the historically merged source models. Omitting the source index $i$ for clarity, we extract all model update vectors:
\begin{equation} \label{eq:muv}
\{\mathbf{v}^l\}_{l=1}^{N_L} =\{\theta^l-\theta^l_0\}_{l=1}^{N_L}, 
\end{equation}
and concatenate them into a flattened vector:
\begin{equation} \label{eq:cat}
{\mathbf{v}_{cat}} = \operatorname{cat}(\mathbf{v}^1, \mathbf{v}^2,... ,\mathbf{v}^{N_L}) \in \mathbb{R}^{1\times\sum_{l=1}^{N_L} D_l},
\end{equation}
where $D_l$ is the dimension of the $l$-th vector.
To sparsify this vector, we zero out redundant parameters using a threshold $\sigma$ computed based on the percentile $r$ of parameter magnitudes:
\begin{equation}\label{eq:sigma}
    \sigma = \operatorname{Percentile}(\operatorname{sort}(\tilde{\mathbf{v}}_{cat}), r).
\end{equation}
Note that for models with a large number of parameters, the dimension of $\tilde{\mathbf{v}}_{cat}$ is very high, and estimating $\sigma$ may cause memory pressure, which can be roughly approximated by random sampling.
We then compute a binary mask $\mathbf{m}$:
\begin{equation}\label{eq:mask}
    \mathbf{m} = \mathbb{I}(\mathbf{v}_{cat}>\sigma),\  \mathbf{m} \in \mathbb{R}^{1\times\sum_{l=1}^L D_l}.
\end{equation}
Applying this mask yields the trimmed model update vector, which we split back into layer-wise vectors:
\begin{equation} \label{eq:split}
\begin{split}
    &\tilde{\mathbf{v}}_{cat} = \mathbf{m}\cdot \mathbf{v}_{cat}, \\ &\tilde{\mathbf{v}}^1,\tilde{\mathbf{v}}^2,...,\tilde{\mathbf{v}}^L = \operatorname{split}(\tilde{\mathbf{v}}_{cat}).
\end{split}
\end{equation}
For each vector obtained from the source model, we calculate the average of all non-zero values to obtain a trimmed average vector, and add the retained updates back to the original CLIP model. 
We omit the layer index $l$, and formulate this process as,
\begin{equation} \label{eq:muvmerge}
\begin{split}
    &\mathbf{v}_{merge} = \frac{\sum_{i=1}^{N_S} \tilde{\mathbf{v}}_i}{\sum_{i=1}^{N_S} \mathbf{m}_i}\cdot \mathbb{I}(\sum_{i=1}^{N_S} \mathbf{m}_i > 0), \\ 
    &\theta_{final} = \theta_0+\mathbf{v}_{merge}.
\end{split}
\end{equation}
We denote our redundancy-aware merging operation as $\operatorname{RHM}(\cdot)$, which take the source model $\{f_1, f_2,...,f_{N_S}\}$ with corresponding parameters $\{\theta_1, \theta_2,...,\theta_{N_S}\}$ and a original CLIP model $f_0$ with parameters $\theta_0$, and outputs a final merged model $f_{final}$ with parameters $\theta_{final}$.
The proposed overall model merging methods in Eq. (\ref{eq:merge1}) can be reformulated as,
\begin{equation}\label{eq:merge2}
\begin{split}
    &\theta_{final} = \operatorname{RHM}(\theta_0, \theta_1, \theta_2,...,\theta_{N_S}),\\
    &\theta_{i} = \sum_{t=0}^T\frac{\gamma_t}{\sum_{k=0}^T\gamma_k}\cdot\theta_{i,t}, \quad i=1,2,...,N_S.
\end{split}
\end{equation}
Some research \citep{yadav2023ties} trim parameters by using Eq. (\ref{eq:sigma}) to estimate thresholds at each layer during merging. 
Our method trims overall parameters. 
This ensures that we can perceive global redundancy.

The overall objective of HAM for each source model can be formulated as:
\begin{equation}\label{eq:all}
    \mathcal{L} = \mathcal{L}_{CE} + \lambda\mathcal{L}_{sign},
\end{equation}
where $\lambda$ is a hyperparameter.
Each model is optimized under the constraint of $\mathcal{L}$, harmonizing the parameters of the source models while learning domain-specific knowledge.

\noindent
\textbf{Algorithm.}
In Algorithm \ref{alg}, we provided the overall process of HAM, where the three modules of our method are abbreviated as SAE (Sample Conflict-aware Adaptive Source Enrichment), OPA (Optimization Conflict-aware Parameter Alignment), and RHM (Redundancy-aware Historical Model Merging), clearly indicating how they are executed in the algorithm.

\begin{algorithm}
\caption{Procedure of HAM.}
\label{alg}
\SetAlgoLined

\KwIn{
    Total number of source domains $N_S$; total number of the layer of vision encoders $N_L$; Source datasets $\{\mathcal{X}_i\}_{i=1}^{N_S}$; A frozen shared CLIP text encoder $\xi$;  Source visual encoder parameters $\{\theta_i\}_{i=1}^{N_S}$ initialized by CLIP's $\theta_0$; Loss weight $\lambda$; Trimming ratio $r$; Total number of training steps $N_T$.
}
\KwOut{Merged model $\theta_{final}$}

\textbf{Training Stage:} \\
\ForEach{step $t \in \{1,\ldots,N_T\}$}{
    
    Calculate model update vectors: $\{\mathbf{v}_{i,t} = \theta_{i,t} - \theta_0\}_{i=1}^{N_S}$ \\
    Calculate the average: 
    $\bar{\mathbf{v}}_t = \frac{1}{N_S}\sum_{i=1}^{N_S}\mathbf{v}_{i,t}$ \\
    Detach the gradient of $\bar{\mathbf{v}}_t$ \\

    \ForEach{source $ i \in \{1,\ldots,N_S\}$}{
    Sample mini-batches $B_i$ from $\mathcal{X}_i$ \\
    (\textbf{SAE}) Calculate the adaptive threshold $\tau_i$ by Eq. (\ref{eq:tau}) \\
    (\textbf{SAE}) Obtain enriched source batch $B'_i$ by Eq. (\ref{eq:enrich}) \\
    (\textbf{OPA}) Calculate the sign regularization by Eq. (\ref{eq:sign}) \\
    (\textbf{OPA}) Update $\theta_{i,t}$ by the overall objective Eq. (\ref{eq:all}) \\
    (\textbf{RHM}) Aggregate historical models along the training trajectory by Eq. (\ref{eq:bma})
    }
}

\textbf{Merging Stage:} \\
\ForEach{source $ i \in \{1,\ldots,N_S\}$}{
Extract the concatenated model update vectors of $\mathbf{v}_{cat}$ by Eq. (\ref{eq:cat}) \\
(\textbf{RHM}) Calculate the mask $\mathbf{m}$ with $r$ by Eq. (\ref{eq:mask}) \\
(\textbf{RHM}) Trim and split the $\mathbf{v}_{cat}$ to obtain $\{\tilde{\mathbf{v}}^l\}_{l=1}^{N_L}$ by Eq. (\ref{eq:split})\\
}

Obtain final merged model update vector $\mathbf{v}_{merge}$ by Eq. (\ref{eq:muvmerge}) \\
Updates back to the original CLIP $\theta_{final} = \theta_0+\mathbf{v}_{merge}$

\Return $\theta_{final}$ \\
\end{algorithm}

\section{Experiments} \label{sec: exp}
This section demonstrates the superiority of our method across five widely used domain generalization benchmark datasets.
Moreover, we perform detailed ablation studies to evaluate the impact of our source data enrichment strategy, parameter alignment strategy (\ie, sign loss $\mathcal{L}_{dir}$), and historical model merging strategy.
In addition, we present sensitivity analyses and qualitative results.

\subsection{Datasets and implementation details}
We conducted experiments on five widely used domain generalization benchmark datasets, encompassing a diverse range of characteristics, including dataset size (small- to large-scale), class diversity (few to many classes), and image type (drawing to real-world).
Next, we provide a detailed introduction to each dataset.

\noindent
\textbf{PACS} \citep{li2017deeper} encompasses four distinct domains: Photo (P), Art Painting (A), Cartoon (C), and Sketch (S). 
It includes seven object classes (dog, elephant, giraffe, guitar, horse, house, and person).
The dataset contains approximately 9,991 images, with significant domain imbalance (e.g., Photo has the most samples). 
The challenge lies in the visual style differences between domains, such as adapting from synthetic Cartoon images to real-world Photos. 

\noindent
\textbf{VLCS} \citep{torralba2011unbiased} integrates four classic computer vision datasets to establish a benchmark for domain generalization. 
It encompasses 5 common object categories and contains approximately 10,729 images, with each domain contributing roughly 2,000–3,000 samples. 
The domains exhibit variations in image capture conditions and scene compositions, rendering VLCS particularly suitable for evaluating models' robustness to small-scale domain shifts. 
Due to its simplicity and balanced domain distribution, VLCS has been widely employed in early domain adaptation and generalization research.

\noindent
\textbf{OfficeHome} \citep{venkateswara2017deep} focuses on indoor scenes, with 65 fine-grained categories of office and household objects (\eg, keyboard, notebook, vase). 
It spans 4 domains: Art, ClipArt, Product, and Real, totaling about 15,588 images. 
The domains vary in image sources, such as product photos versus hand-drawn art, creating significant domain shifts. 
The fine-grained classification and large number of categories make OfficeHome a challenging benchmark for evaluating models' scalability and generalization across complex domains.

\noindent
\textbf{TerraIncognita (TerraInc)} \citep{beery2018recognition} is a wildlife dataset collected from camera traps across four distinct geographic locations. 
It comprises 10 animal species (\eg, leopard, hyena, rhino) and contains approximately 24,788 images, exhibiting some class imbalance (\eg, a high proportion of empty background images). 
The domain shifts arise from variations in environmental conditions, such as lighting and occlusions, across locations. 
This dataset is particularly well-suited for evaluating models' robustness in real-world, long-tailed distribution scenarios and their ability to generalize to unseen ecological environments.

\noindent
\textbf{DomainNet (DN)} \citep{peng2019moment} is the largest domain generalization dataset, encompassing 345 fine-grained categories of everyday objects (\eg, balloon, map) across six distinct domains: Real, Sketch, ClipArt, Painting, Quickdraw, and Text. 
It comprises approximately 600,000 images, exhibiting significant domain heterogeneity, such as the contrast between real photos and abstract sketches. 
The dataset is particularly challenging due to its large-scale domain shifts and the presence of label noise in certain categories. 
DomainNet is widely utilized to evaluate the generalization capabilities of large-scale pre-trained models across highly diverse and complex domains.

Collectively, these datasets provide comprehensive benchmarks for evaluating domain generalization algorithms, encompassing a wide range of domain shifts—from variations in visual styles to ecological differences and cross-modal contrasts.
Following the training and evaluation protocol of DomainBed \citep{gulrajani2020search}, we select one domain for
testing and the remaining domains for training every time,
and 20\% samples of the training domains are held out for validation and model selection. 
Model selection is carried out based on the training-domain validation set.

\noindent
\textbf{Implementation Details.}
Through a random hyperparameter search, we determine learning rates, total training steps, the trimming threshold $\tau$, and the weight $\lambda$ of direction loss. 
The search space for the learning rate is \{5e-6, 1e-5\}, for the total training steps is \{500, 3000, 5000, 10000\}, for the trimming ratio $r$ is \{0.2, 0.4, 0.6, 0.8\}, and for the weight $\lambda$ is \{0.1, 0.5, 1.0\}.
We utilize the AdamW optimizer \citep{loshchilov2017decoupled} and set a batch size of 24 and a weight decay of 0.1 for all experiments.
All results are the average of seeds \{41, 42, 43\}.
The experiments are conducted on NVIDIA GeForce 3090 GPUs with 24GB of memory, using the PyTorch framework as the deep learning architecture.

\subsection{Main Results}

\noindent
\textbf{Baselines and Compared Methods.}
We evaluate HAM against the state-of-the-art (SOTA) approaches on the five standard benchmark datasets.
Then we introduce these methods.
ZS-CLIP \citep{radford2021learning} is the zero-shot classification results of the original CLIP, serves as a pre-trained vision-language baseline model without any training.
Empirical risk minimization (ERM) is the traditional DG method, which minimizes average training loss.
ZS-CLIP and ERM indicate the baseline performance without and with training, respectively.
Besides, we benchmark our approach against a wide range of domain generalization strategies, 
including 1) invariant learning methods, IRM \citep{arjovsky2019invariant}, IIB \citep{li2022invariant}, MIRO \citep{cha2022domain},  CLIPCEIL++ \citep{yu2024clipceil}, learns domain-invariant information through various techniques;
2) distribution alignment methods, MMD \citep{li2018domain}, aligns domain distributions via kernel-based divergence;
4) prompt-generation method, SPG \citep{bai2024soft}, generate task-specific prompts for unseen target domains.
3) robust learning methods, CLIPood \citep{shu2023clipood}, CAR-FT \citep{mao2024context}, VLV2-SD \citep{addepalli2024leveraging}, AlignCLIP \citep{han2025alignclip}, SFT \citep{li2024seeking}, aims to improve the model robustness against distribution shifts.
As existing methods predominantly adopt the ViT-B/16 \citep{dosovitskiy2020image} backbone, we focus our comparisons on ViT-B/16-based approaches. 
To validate the generalization performance of HAM, we further report results for representative state-of-the-art (SOTA) methods adapted to the ResNet50 \citep{he2016deep} architecture. 
Results marked with an asterisk (*) were reproduced in our experiments; 
all others are sourced directly from the original CLIP-CEIL \citep{yu2024clipceil} and SFT \citep{li2024seeking} publications.

\noindent
\textbf{Experimental Analysis.}
Table \ref{exp:clip} presents a comprehensive comparison of our proposed HAM method against state-of-the-art (SOTA) domain generalization approaches across five widely-used benchmark datasets (PACS, VLCS, OfficeHome, TerraInc, and DomainNet). 
The evaluation is conducted on two backbone architectures, ResNet50 and ViT-B/16.
For the ResNet50 backbone, HAM achieves an average accuracy of 70.9\%, outperforming all baseline methods. 
Compared to the zero-shot CLIP (ZS-CLIP) baseline (62.5\%), HAM improves by 8.4 percentage points, demonstrating the necessity of domain adaptation for generalization. 
Moreover, HAM surpasses the empirical risk minimization (ERM) approach by 4.1\%, highlighting its superiority over naive supervised training. 
Notably, compared to previous robust learning methods such as CLIPood (70.6\%) and CLIPCEIL++ (70.7\%), HAM still achieves the highest performance, validating its effectiveness in enhancing robustness under domain shifts.
Breaking down the results by datasets, HAM exhibits particularly strong performance on PACS (93.7\%), VLCS (82.7\%), and DomainNet (52.7\%), showing its ability to generalize across diverse domains.
For the ViT-B/16 backbone, HAM again achieves the best overall performance, with an average accuracy of 79.0\%, surpassing all competing methods. 
Compared to ZS-CLIP (70.3\%), HAM improves by 8.7\%, while also outperforming ERM (74.3\%) by 4.7\%. 
This demonstrates that even with a stronger backbone, domain generalization techniques remain essential for improving generalization.
HAM outperforms invariant learning methods such as IRM (75.8\%) and IIB (75.8\%), as well as distribution alignment approaches like MMD (74.6\%), confirming that a simple alignment of feature distributions is insufficient for robust domain adaptation. 
Compared to advanced robust learning techniques, including CLIPood (78.6\%), CAR-FT (78.5\%), and AlignCLIP (78.8\%), HAM achieves state-of-the-art performance. 
Specifically, it excels in the PACS, VLCS, and TerraInc datasets, where domain shifts are more severe, demonstrating its robustness to distributional changes.

\begin{table*}[ht]
\caption{Comparisons of our methods with the State-of-the-Art methods across five widely-used domain generalization benchmark datasets. The {\bst red} denotes the best, and {\scd green} denotes the second-best results. Results denoted by a star (*) are reproduced by us, and the rest are from CLIPCEIL++ \citep{yu2024clipceil} and SFT \citep{li2024seeking}.}
\label{exp:clip}
\centering
\resizebox{0.9\textwidth}{!}{
\begin{tabular}{l|c|ccccc>{\columncolor[HTML]{EFEFEF}}c}
\toprule
Method & Arch & PACS & VLCS & OfficeHome & TerraInc & DomainNet & Avg. \\ \midrule
ZS-CLIP* \citep{radford2021learning} &  & 90.9 & 79.7 & 71.8 & 23.4 & 46.6 & 62.5 \\
ERM* &  & 91.3 & 80.9 & 68.4 & {\scd 49.2} & 44.2 & 66.8 \\
SPG \citep{li2024seeking} &  & 92.8 & {\bst 84.0} & 73.8 & 37.5 & 50.1 & 67.6 \\
CLIPood* \citep{shu2023clipood} &  & {\scd 93.0} & {\scd 82.9} & {\bst 76.6} & 47.6 & {\bst 52.7} & {\scd 70.6} \\
CLIPCEIL++* \citep{yu2024clipceil} &  & 92.6 & 82.4 & 75.7 & 47.7 & 51.2 & 69.9 \\
Ours & \multirow{-6}{*}{\begin{tabular}[c]{@{}c@{}}CLIP\\ (RN50)\end{tabular}} & {\bst 93.7} & 82.7 & {\scd 75.9} & {\bst 49.3} & {\bst 52.7} & {\bst 70.9} \\ \midrule
ZS-CLIP* \citep{radford2021learning} &  & 96.2 & 81.8 & 82.0 & 33.8 & 57.5 & 70.3 \\
ERM* &  & 95.6 & 82.2 & 83.2 & 59.7 & 57.3 & 75.6 \\
IRM \citep{arjovsky2019invariant} &  & 96.4 & 81.9 & 83.1 & 50.9 & 59.1 & 74.3 \\
MMD \citep{li2018domain} &  & 95.1 & 81.9 & 83.7 & 56.9 & 59.9 & 75.5 \\
IIB \citep{li2022invariant} &  & 96.0 & {\scd 82.5} & 83.9 & 58 & 58.6 & 75.8 \\
MIRO \citep{cha2022domain} &  & 95.6 & 82.2 & 82.5 & 54.3 & 54.0 & 73.7 \\
CLIPood* \citep{shu2023clipood} &  & 97.1 & 84.7 & 86.7 & 59.0 & {\scd 63.4} & {\scd 78.6} \\
CAR-FT \citep{mao2024context} &  & 96.8 & {\scd 85.5} & 85.7 & {\bst 61.9} & 62.5 & 78.5 \\
SPG \citep{bai2024soft} &  & 97.0 & 82.4 & 83.6 & 50.2 & 60.1 & 74.7 \\
VLV2-SD \citep{addepalli2024leveraging} &  & 96.7 & 83.3 & {\bst 87.4} & 58.5 & 62.8 & 77.7 \\
CLIPCEIL++* \citep{yu2024clipceil} &  & 97.1 & 83.8 & 85.7 & 58.4 & 62.8 & 77.6 \\
AlignCLIP \citep{han2025alignclip} & \multirow{-12}{*}{\begin{tabular}[c]{@{}c@{}}CLIP\\ (ViT-B/16)\end{tabular}} & {\scd 97.3} & 85.1 & 86.9 & 59.5 & {\bst 63.5} & 78.5 \\
SFT \citep{li2024seeking} &  & 96.8 & 84.1 & 86.5 & 61.2 & 60.5 & 77.8 \\
Ours &  & {\bst 97.4} & {\bst 85.8} & {\scd 87.3} & {\scd 61.4} & {\scd 63.4} & {\bst 79.0} \\ \bottomrule
\end{tabular}
}
\end{table*}

\subsection{Ablation Study}
Our method includes three parts: source conflict-aware adaptive source enrichment (SAE), optimization conflict-aware parameter alignments (OPA), and redundancy-aware model merging (RHM). 
SAE is designed to enhance the inherent capabilities of the source model, while OPA shares the same purpose as RHM, which harmonizes the source model to reduce conflicts during the parameter merging process. 
In this section, we conduct ablation experiments on these three modules to explore their effectiveness.

\begin{table*}[ht]
\caption{Ablation study on SAE.} \label{abl:sample}
\centering
\resizebox{0.6\textwidth}{!}{
\begin{tabular}{cccc}
\toprule
Single source only & w/ SAE & Acc on TerraInc & Acc on OfficeHome \\ \midrule
$\checkmark$ & $\times$ & 54.8 & 85.9 \\
$\checkmark$ & $\checkmark$ & 61.4 & 87.3 \\ \bottomrule
\end{tabular}
}
\end{table*}

\noindent
\textbf{Ablation study on SAE.}
As shown in Table \ref{abl:sample}, the introduced SAE significantly improves the model’s generalization ability. 
By incorporating SAE, the model benefits from a more robust representation, leading to a substantial increase to 61.4\% on TerraInc and 87.3\% on OfficeHome, demonstrating that addressing sample conflicts at the source level is crucial for improving cross-domain generalization.
Samples from other domains without conflicts enhance the source models evidently.

\begin{table*}[ht]
\caption{Ablation Study on OPA and RHM.}\label{abl:merge}
\centering
\resizebox{0.75\textwidth}{!}{
\begin{tabular}{c|ccc|cc}
\toprule
\multirow{2}{*}{OPA (Sign Loss)} & \multicolumn{3}{c|}{Merging} & \multirow{2}{*}{Acc on TerraInc} & \multirow{2}{*}{Acc on OfficeHome} \\ \cmidrule{2-4}
 & Avg-Merging & Ties-Merging & RHM &  &  \\ \midrule
$\times$ & $\checkmark$ & $\times$ & $\times$ & 60.0 & 86.5 \\
$\times$ & $\times$ & $\checkmark$ & $\times$ & 60.2 & 86.6 \\
$\times$ & $\times$ & $\times$ & $\checkmark$ & 60.0 & 86.7 \\
$\checkmark$ & $\checkmark$ & $\times$ & $\times$ & 60.1 & 86.5 \\
$\checkmark$ & $\times$ & $\checkmark$ & $\times$ & 60.6 & 86.8 \\
$\checkmark$ & $\times$ & $\times$ & $\checkmark$ & 61.4 & 87.3 \\ \bottomrule
\end{tabular}
}
\end{table*}

\noindent
\textbf{Ablation study on OPA and RHM.}
OPA and RHM are proposed to harmonize and merge the source models, with a regularization term, \ie, sign loss $\mathcal{L}_{sign}$, and a merging strategy.
These two modules operate synergistically.
OPA ensures consistent model update directions and mitigates sign conflicts during parameter merging.
RHM incorporates training trajectory information to smooth parameter updates while eliminating interference from redundant parameters.
Besides, we additionally compare our method with an off-the-shelf merging technique, Ties-Merging \citep{yadav2023ties}, which incorporates sign electing and layer-level redundancy trimming at the merging stage.
Average merging (Avg-Merging) directly takes the average of model update vectors without any redundancy trimming.
Table \ref{abl:merge} presents a detailed analysis, demonstrating that both modules contribute to improved generalization. 
Employing RHM alone increases accuracy to 60.0\% on TerraInc and 86.7\% on OfficeHome, confirming that redundancy-aware merging mitigates adverse effects from parameter conflicts. 
OPA, by enforcing sign consistency in parameter updates, further stabilizes the merging process, particularly when combined with the Ties-Merging strategy, leading to accuracy improvements of 60.6\% on TerraInc and 86.8\% on OfficeHome. 
The best performance is achieved when both OPA and RHM are applied jointly, reaching 61.4\% on TerraInc and 87.3\% on OfficeHome, demonstrating their complementary effects. 
OPA ensures parameter alignment, while RHM incorporates training trajectory information to smooth parameter updates, thereby reducing optimization conflicts and preserving crucial knowledge from individual source models.

\begin{table}[ht]
\centering
\caption{Ablation study on parameter trimming strategy.} \label{abl:trim}
\resizebox{0.4\textwidth}{!}{
\begin{tabular}{ccc}
\toprule
  & Acc on TerraInc & Acc on OfficeHome \\ \midrule
No trimming & 60.1 & 86.5  \\
Layer-level & 60.2 & 86.8 \\
Model-level & 61.4 & 87.3 \\ \bottomrule
\end{tabular}
}
\end{table}

\noindent
\textbf{Ablation study on trimming strategy in RHM.}
Additionally, we explore the impact of parameter trimming in RHM to understand its role in refining the merged model. 
Table \ref{abl:trim} compares different pruning strategies, revealing that without parameter trimming, accuracy remains at 60.1\% on TerraInc and 86.5\% on OfficeHome, indicating that direct parameter merging introduces redundant information that limits generalization. 
Layer-level pruning provides minor improvements, achieving 60.2\% on TerraInc and 86.8\% on OfficeHome, while model-level pruning leads to the most significant gains, reaching 61.4\% on TerraInc and 87.3\% on OfficeHome. 
These findings emphasize that eliminating redundant parameters is crucial for improving model compactness and stability, with model-level pruning proving to be the most effective strategy.

\noindent
\textbf{Ablation study on historical model merging in RHM.}
We compare the results of the best model merging and our historical model merging in Table \ref{abl:bma}.
The best model denotes the model with the highest accuracy on the validation set during training.
Our historical model merging strategy outperforms the best model merging in both the TerraInc and OfficeHome datasets.
This demonstrates that aggregating models across training iterations can enhance robustness and generalization, due to better optimization trajectory exploration and reduced overfitting.

\begin{table}
\caption{Ablation study on historical model merging.}\label{abl:bma}
\centering
\resizebox{0.45\textwidth}{!}{
\begin{tabular}{ccc}
\toprule
 & Acc on TerraInc & Acc on OfficeHome \\ \midrule
Best model & 60.8 & 86.5 \\
Historical model & 61.4 & 87.3 \\ \bottomrule
\end{tabular}
}
\end{table}


Overall, our ablation studies provide strong empirical evidence for the effectiveness of each proposed component. 
SAE plays a pivotal role in resolving sample conflicts at the source level, leading to more robust and transferable representations. 
OPA and RHM work synergistically to ensure a smooth merging process, minimizing optimization conflicts and parameter redundancy. 
Furthermore, parameter trimming refines the merged model by removing unnecessary parameters, with model-level trimming yielding the most substantial improvements. 
These results collectively demonstrate the efficacy of our approach in enhancing cross-domain generalization while maintaining stable and efficient model merging.


\subsection{Sensitivity Analysis}
\begin{figure*}
    \centering
    \includegraphics[width=1\linewidth]{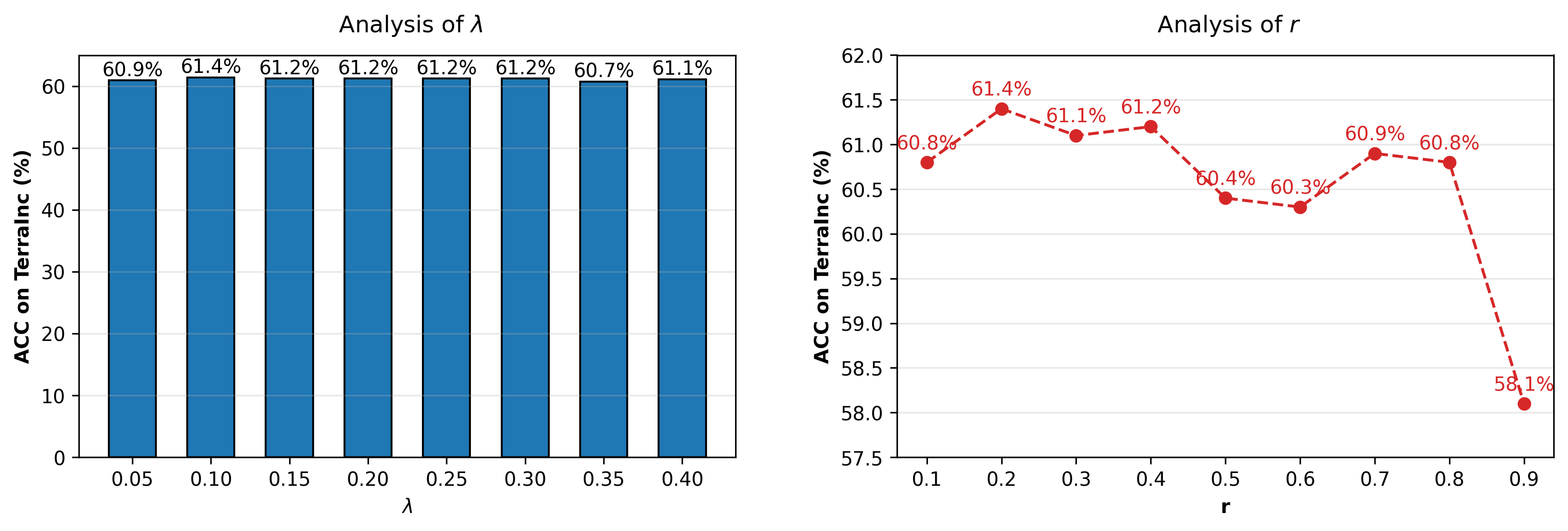}
    \caption{Sensitivity analysis on TerraInc of the weight $\lambda$ of sign loss in Eq. (\ref{eq:all}), and redundancy parameter trimming ratio $r$ in Eq. (\ref{eq:sigma}).}
    \label{fig:sensi}
\end{figure*}

To further investigate the stability and robustness of our approach, we conduct a sensitivity analysis on two key hyperparameters: the regularization weight $\lambda$ in Eq. (\ref{eq:all}) and the redundancy trimming ratio $r$ in Eq. (\ref{eq:sigma}). 
Fig. \ref{fig:sensi} presents the impact of varying these parameters on model performance, measured by average accuracy across four domains in the TerraInc \citep{beery2018recognition} dataset.

For $\lambda$, which controls the strength of the sign consistency regularization in OPA, we observe in the left subfigure that model performance remains relatively stable across a wide range of values. The accuracy fluctuates slightly around 61.2\%–61.4\%, peaking at $\lambda$ = 0.1 with an accuracy of 61.4\%. 
This indicates that the sign consistency loss effectively regularizes parameter updates without being overly sensitive to its weighting, reinforcing its robustness in aligning optimization directions.
For $r$, which determines the proportion of redundant parameters trimmed in RHM, the right subfigure reveals a more pronounced effect on performance. 
When $r$ is small ($\leq$0.2), accuracy remains stable, reaching a maximum of 61.4\% at $r$ = 0.2. 
However, as $r$ increases beyond 0.5, accuracy begins to decline, dropping significantly to 58.1\% at $r$ = 0.9. 
This suggests that moderate trimming effectively removes redundant parameters while preserving essential model capacity, whereas excessive pruning leads to the loss of critical information, ultimately degrading generalization performance.

These findings highlight the robustness of our method to $\lambda$ variations and the importance of carefully selecting $r$ to balance redundancy removal and model expressiveness.
The results validate our design choices, demonstrating that the proposed regularization and trimming strategies contribute to stable and efficient model merging while ensuring strong generalization across unknown domains.

\subsection{Qualitative Analysis}

\begin{figure*}
    \centering
    \includegraphics[width=1\linewidth]{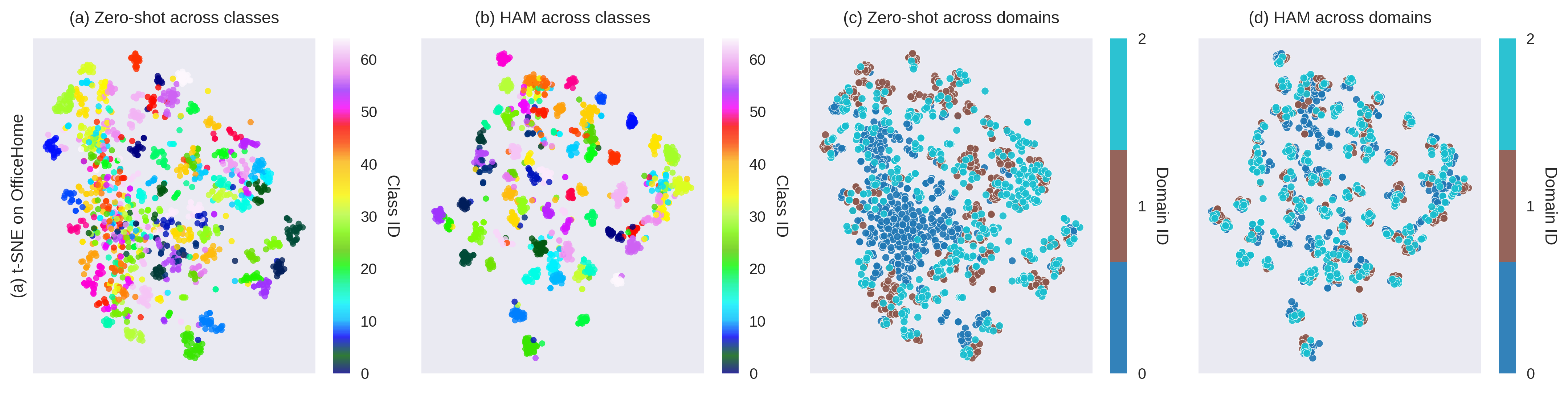} \\
    \includegraphics[width=1\linewidth]{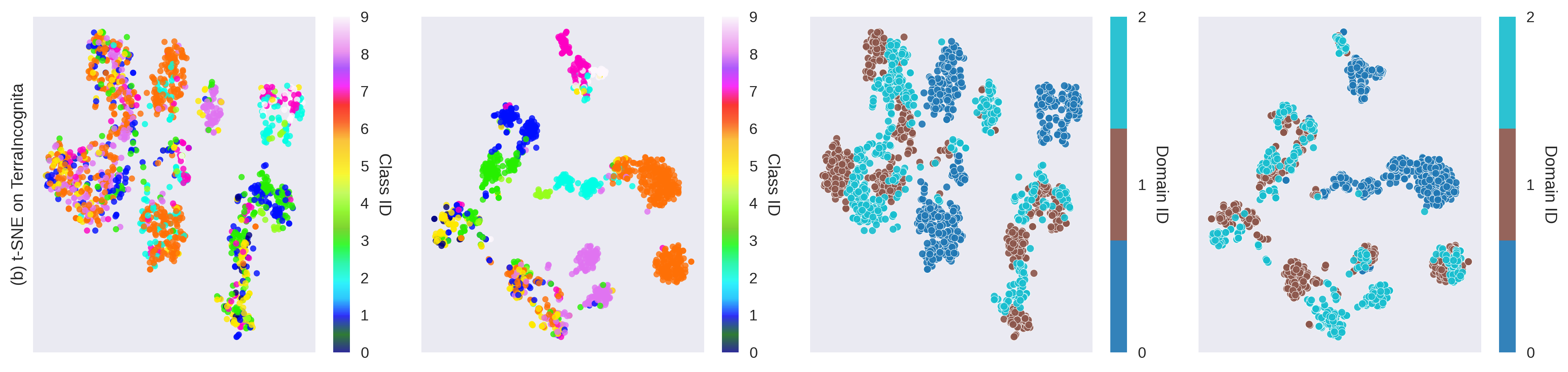}
    \caption{t-SNE \citep{van2008visualizing} visualization on feature points of HAM and CLIP pre-trained models across different classes and domains. Different colors indicate different classes or domains.}
    \label{fig:tsne}
\end{figure*}

To gain deeper insights into the feature representations learned by our method, we perform t-SNE \citep{van2008visualizing} visualizations comparing the feature distributions of our model with the initial CLIP model across different classes and domains. 
Fig. \ref{fig:tsne} presents these comparisons, where different colors denote different class or domain IDs.

In the class-wise analysis, the zero-shot CLIP model exhibits significant class overlap and dispersion, indicating suboptimal class separation. 
In contrast, our model achieves more compact and well-clustered class distributions, demonstrating its effectiveness in improving feature discrimination across classes. 
This enhanced clustering suggests that HAM better aligns visual representations with their corresponding semantic categories, leading to improved classification performance.
For the domain-wise analysis, the features across domains of zero-shot CLIP are dispersed.
This suggests that CLIP’s pretrained features may not be sufficiently domain-invariant, limiting its generalization across unseen distributions. 
With HAM, the learned representations exhibit better domain alignment, implying that our method effectively reduces domain shifts and improves cross-domain robustness.


\subsection{Results beyond CLIP models}

\begin{table*}[ht] 
\caption{Comparisons of our methods with the State-of-the-Art methods across five domain generalization benchmark datasets on \textbf{ImageNet pre-trained ResNet-50} backbone. Results are from SFT \citep{li2024seeking} and the corresponding papers except ours.}\label{exp:rn50}
\centering
\resizebox{0.8\textwidth}{!}{
\begin{tabular}{l|c|ccccc>{\columncolor[HTML]{EFEFEF}}c}
\toprule
Method & Arch & PACS & VLCS & OfficeHome & TerraInc & DomainNet & Avg. \\ \midrule
ERM &  & 85.5 & 77.5 & 66.5 & 46.1 & 40.9 & 63.3 \\
IRM \citep{arjovsky2019invariant} &  & 83.5 & 78.5 & 64.3 & 47.6 & 33.9 & 61.6 \\
Mixup \citep{zhang2017mixup} &  & 84.6 & 77.4 & 68.1 & 47.9 & 39.2 & 63.4 \\
CORAL \citep{sun2016deep} &  & 86.2 & 78.8 & 68.7 & 47.6 & 41.5 & 64.6 \\
MMD \citep{li2018domain} &  & 84.6 & 77.5 & 66.3 & 42.2 & 23.4 & 58.8 \\
SagNet \citep{nam2021reducing} &  & {\scd 86.3} & 77.8 & 68.1 & 48.6 & 40.3 & 64.2 \\
Fish \citep{shi2022gradient} &  & 85.5 & 77.8 & 68.6 & 45.1 & 42.7 & 63.9 \\
SelfReg \citep{kim2021selfreg} &  & 85.6 & 77.8 & 67.9 & 47.0 & 42.8 & 64.2 \\
mDSDI \citep{bui2021exploiting} &  & 86.2 & {\scd 79.0} & 69.2 & 48.1 & 42.8 & 65.1 \\
MIRO \citep{cha2022domain}&  & 85.4 & {\scd 79.0} & 70.5 & 50.4 & 44.3 & 65.9 \\
Mixstyle \cite{zhou2024mixstyle} &  & 85.2 & 77.9 & 60.4 & 44.0 & 34.0 & 60.3 \\
Hutchinson \cite{hemati2023understanding} &  & 83.9 & 76.8 & 68.2 & 46.6 & 41.6 & 63.4 \\
Arith \cite{wang2025balanced} &  & 86.5 & 79.4 & 69.4 & 48.1 & 41.5 & 65.0 \\
LFME \cite{chen2024lfme} &  & 85.0 & 78.4 & 69.1 & 48.3 & 42.1 & 64.6 \\
I$^3$C \cite{zhou2025learning} &  & {\scd 87.1} & {\scd 79.6} & {\scd 70.2} & 49.6 & 42.6 & 65.8 \\
SFT \cite{li2024seeking} &  & {\bst 88.3} & {\bst 79.8} & {\bst 70.9} & {\bst 50.7} & {\scd 46.0} & {\bst 67.1} \\
Ours & \multirow{-17}{*}{RN50} & 86.2 & 78.7 & 69.6 & {\scd 50.5} & {\bst 46.9} & {\scd 66.4}

\\ \bottomrule
\end{tabular}
}
\end{table*}

To validate the generality of HAM on non-CLIP models, we further conduct experiments on ResNet-50 pre-trained on ImageNet \citep{deng2009imagenet}.
We compare HAM with several classical methods, focusing on traditional domain generalization: Mixup \citep{zhang2017mixup}, CORAL \citep{sun2016deep}, MMD \citep{li2018domain}, SagNet \citep{nam2021reducing}, Fish \citep{shi2022gradient}, SelfReg \citep{kim2021selfreg}, mDSDI \citep{bui2021exploiting}, MIRO \citep{cha2022domain}, Hutchinson \cite{hemati2023understanding}, Arith \cite{wang2025balanced}, LFME \cite{chen2024lfme}, I$^3$C \cite{zhou2025learning}, and SFT \citep{li2024seeking}.

As shown in Table \ref{exp:rn50}, our method achieves a competitive average accuracy of 66.4\%, demonstrating the effectiveness of HAM beyond CLIP backbones. This result confirms that our approach is not limited to vision-language models but also generalizes well to standard ImageNet-pretrained ResNet-50 architectures.
Compared to strong DG baselines such as ERM (63.3\%), IRM (61.6\%), and CORAL (64.6\%), our method consistently outperforms them, highlighting the advantage of explicitly addressing sample and optimization conflicts in domain generalization. 
Additionally, our approach achieves comparable or superior performance to advanced methods like I$^3$C (65.8\%) and LFME (64.6\%), further demonstrating the robustness of HAM in handling diverse domain shifts.
Notably, although SFT (67.1\%) achieves the highest average accuracy, our method remains highly competitive while maintaining a lighter and more efficient framework without requiring additional self-training mechanisms. 
The strong performance of HAM across all datasets validates that model merging is a promising strategy for domain generalization, extending beyond CLIP-based architectures to traditional convolutional backbones.

\section{Conclusions} \label{sec: con}

In this work, we reveal two inherent conflicts in the traditional single-model optimization paradigm: sample conflict and optimization conflict, which arise due to the varying quality of datasets from different domains and the inherent trade-offs in multi-objective optimization. 
To address these challenges, we propose Harmonizing and Merging (HAM), a novel approach that first trains domain-specific models and then merges them into a unified model.
HAM consists of three modules termed sample conflict-aware adaptive source enrichment, optimization conflict-aware parameter alignment, and redundancy-aware historical model merging.
By systematically resolving these conflicts through our three proposed modules, HAM effectively consolidates domain-specific knowledge while mitigating the interference caused by noisy samples and divergent optimization trajectories. 
Unlike ensemble learning, which requires storing multiple models, HAM merges models at the parameter level, significantly reducing storage overhead while preserving generalization capacity.
To the best of our knowledge, HAM is the first attempt at model merging in the domain generalization field. 
The strong performance of HAM demonstrates the great potential of model merging for DG.

\section*{Data Availability Statement}

The datasets in our study can be downloaded in the open-source GitHub repository, \url{https://github.com/facebookresearch/DomainBed}.


\small
{
\bibliographystyle{unsrt}
\bibliography{ref}
}


\end{sloppypar}

\end{document}